# Multiset-Valued Linear Index Grammars: Imposing Dominance Constraints on Derivations

Owen Rambow
Univesité Paris 7
UFR Linguistique, TALANA
Case 7003, 2, Place Jussieu
F-75251 Paris Cedex 05, France
rambow@linguist.jussieu.fr

## Abstract

This paper defines multiset-valued linear index grammar and unordered vector grammar with dominance links. The former models certain uses of multiset-valued feature structures in unification-based formalisms, while the latter is motivated by word order variation and by "quasi-trees", a generalization of trees. The two formalisms are weakly equivalent, and an important subset is at most context-sensitive and polynomially parsable.

## Introduction

Early attempts to use context-free grammars (CFGs) as a mathematical model for natural language syntax have largely been abandoned; it has been shown that (under standard assumptions concerning the recursive nature of clausal embedding) the cross-serial dependencies found in Swiss German cannot be generated by a CFG (Shieber, 1985). Several mathematical models have been proposed which extend the formal power of CFGs, while still maintaining the formal properties that make CFGs attractive formalisms for formal and computational linguists, in particular, polynomial parsability and restricted weak generative capacity. These mathematical models include tree adjoining grammar (TAG) (Joshi et al., 1975; Joshi, 1985), head grammar (Pollard, 1984), combinatory categorial grammar (CCG) (Steedman, 1985), and linear index grammar (LIG) (Gazdar, 1988). These formalisms have been shown to be weakly equivalent to each other (Vijay-Shanker et al., 1987; Vijay-Shanker and Weir, 1994); we will refer to them as "LIG-equivalent formalisms". LIG is a variant of index grammar (IG) (Aho, 1968). Like CFG, IG is a context-free string rewriting system, except that the nonterminal symbols in a CFG are augmented with stacks of index symbols. The rewrite rules push or pop indices from the index stack. In an IG, the index stack is copied to all nonterminal symbols on the right-hand side of a rule. In a LIG, the stack is copied to exactly one right-hand side nonterminal.[1]

While LIG-equivalent formalisms have been shown to provide adequate formal power for a wide range of linguistic phenomena (including the aforementioned Swiss German construction), the need for other mathematical formalisms has arisen in several unrelated areas. In this paper, we discuss three such cases. First, capturing several semantic and syntactic issues in unification-based formalisms leads to the use of multiset-valued feature structures. Second, word order facts from languages such as German, Russian, or Turkish cannot be derived by LIG-equivalent formalisms. Third, a generalization of trees to "quasi-trees" (Vijay-Shanker, 1992) in the spirit of D-Theory (Marcus et al., 1983) leads to the definition of a new formal system. In this paper, we introduce two new equivalent mathematical formalisms which provide adequate descriptions for these three phenomena.

The paper is structured as follows. First, we present the three phenomena in more detail. We then introduce multiset-valued LIG and present some formal properties. Thereafter, we introduce a second rewriting system and show that it is weakly equivalent to the LIG variant. We then briefly mention some related formalisms. We conclude with a brief summary.

## Three Problems for LIG-Equivalent Formalisms

The three problems we present are of a rather different nature. The first arises from the way a linguistic problem is treated in a specific type of framework (unification-based formalisms). The second problem derives directly from linguistic data. The third problem is a formalism which has been motivated on independent, methodological grounds, but whose formal properties are unknown.

### Multiset-Valued Feature Structures

HPSG (Pollard and Sag, 1987; Pollard and Sag, 1994) uses typed feature structures as its formal basis, which are Turing-equivalent. However, it is not necessarily the case that the full power of the system is used in the linguistic analyses that are expressed in it. HPSG analyses include information about constituent structure which can be represented as a context-free phrase-structure tree. In addition, various mechanisms have been proposed to handle certain linguistic phenomena that relate two nodes within this tree. One of these is a multiset-valued feature that is passed along the phrase-structure tree from daughter node to mother node. Multiset-valued features have been proposed for the SLASH feature which handles *wh*-dependencies (Pollard and Sag, 1994, Chapter 4), and for certain semantic purposes, including the representation of stored quantifiers in a mechanism similar to Cooper-storage. Another use may be the representation of anti-coreference constraints arising from Principle C of Binding Theory (be it that of (Chomsky, 1981) or of Pollard and Sag (1992)).

---

[1]Note that a LIG is not an IG that is linear (i.e., whose productions have at most one nonterminal on the right-hand side), but rather, it is a context-free grammar with linear indices (i.e., the indices are never copied).

It is desirable to be able to assess the formal power of such a system, for both theoretical and practical reasons. Theoretically, it would be interesting if it turned out that the linguistic principles formulated in HPSG naturally lead to certain restricted uses of the unification-based formalism. Clearly this would represent an important insight into the nature of grammatical competence. On the practical side, formal equivalences can guide the building of applications such as parsers for existing HPSG grammars. For example, it has been proposed that HPSG grammars can be "compiled" into TAGs in order to obtain a computationally more tractable system (Kasper, 1992), thus sidestepping the issue of building parsers for HPSG directly. However, LIG-equivalent formalisms cannot serve as targets for compilations in cases in which HPSG uses multiset-valued feature structures.

## Word Order Variation

Becker et al. (1991) discuss scrambling, which is the permutation of verbal arguments in languages such as German, Korean, Japanese, Hindi, Russian, and Turkish. If there are embedded clauses, scrambling in many languages can affect arguments of more than one verb ("long-distance" scrambling).

(1) ... daß [den Kühlschrank]$_i$   bisher noch
    ... that the refrigerator$_{ACC}$  so far yet
    niemand    [t$_i$ zu reparieren] versprochen hat
    no-one$_{NOM}$  to repair          promised    has

...that so far, no-one has promised to repair the refrigerator

Scrambling in German is "doubly unbounded" in the sense that there is no bound on the number of clause boundaries over which an element can scramble, and an element scrambled (long-distance or not) from one clause does not preclude the scrambling of an element from another clause:

(2) ... daß [dem Kunden]$_i$ [den Kühlschrank]$_j$
    ... that the client$_{DAT}$  the refrigerator$_{ACC}$
    bisher noch niemand    t$_i$ [[t$_j$ zu reparieren]
    so far yet  no-one$_{NOM}$        to repair
    zu versuchen] versprochen hat
    to try        promised    has

...that so-far, no-one yet has promised the client to repair the refrigerator

Similar data has been observed in the literature for other languages, for example for Finnish by Karttunen (1989). Becker et al. (1991) argue that a simple TAG (and the other LIG-equivalent formalisms) cannot derive the full range of scrambled sentences. Rambow and Satta (1994) propose the use of unordered vector grammar (UVG) to model the data. In UVG (Cremers and Mayer, 1973), several context-free string rewriting rules are grouped into vectors, as for *verspricht* 'promises':

(3) $((S \rightarrow NP_{nom} VP), (VP \rightarrow NP_{dat} VP),$
    $(VP \rightarrow S_{inf} V), (V \rightarrow \textit{verspricht})\ )$

During a derivation, rules from a vector can be applied in any order, and rules from different vectors can be interleaved, but at the end, all rules from an instance of a vector must have been used in the derivation. By varying the order in which rules from different vectors are applied, we can derive different word orders. Observe that the vector in (3) contains exactly one terminal symbol (the verb); grammars in which every elementary structure (vector in UVG, tree in TAG, rule in CFG) contains at least one terminal symbol we will call *lexicalized*.

Languages generated by UVG are known to be context-sensitive and semilinear (Cremers and Mayer, 1974) and polynomially parsable (Satta, 1993). However, they are not adequate for modeling natural language syntax. In the following example, (4a) is out since there is no analysis in which the moved NP c-commands its governing verb, as is the case in (4b).

(4) a. * ... daß niemand       [dem Kunden] [t$_i$
       ... that no-one$_{NOM}$   the client$_{DAT}$
       zu versuchen] [den Kühlschrank]$_j$  versprochen
       to try         the refrigerator$_{ACC}$  promised
       hat [t$_j$ zu reparieren]$_i$
       has   to repair

    b. ?  ... daß niemand [dem Kunden] [den
       Kühlschrank]$_j$ [t$_i$ zu versuchen] versprochen hat
       [t$_j$zu reparieren]$_i$

What is needed is an additional mechanism that enforces a dominance relation between the sister node of an argument and its governing verb.

## Quasi-Trees

Vijay-Shanker (1992) introduces "quasi-trees" as a generalization of trees. He starts from the observation that the traditional definition of tree adjoining gram-mar (TAG) is incompatible with a unification-based approach because the trees of a TAG start out as fully specified objects, which are later modified; in particular, immediate dominance relations in a tree need not hold after another tree is adjoined into it. In order to arrive at a definition that is compatible with a unification-based approach, he makes three minimal assumptions about the nature of the objects used for the representation of natural language syntax. The first assumption (left implicit) is that these objects represent phrase-structure. The second assumption is that they "give a sufficiently enlarged domain of locality that allows localization of dependencies such as subcategorization, and filler-gap" (Vijay-Shanker, 1992, p.486). The third assumption is that dominance relations can be stated between different parts of the representation. These assumptions lead Vijay-Shanker to define quasi-trees, which are partial descriptions of trees in which "quasi-nodes" (partial descriptions of nodes) are related by dominance constraints. Each node in a traditional tree (as used in TAG) corresponds to two quasi-nodes, a top and a bottom version, such that the top dominates the bottom.

There are two ways of interpreting quasi-trees: either quasi-trees can be seen as data structures in their own right; or quasi-trees can be seen as descriptions of trees whose denotations are sets of (regular) trees. If quasi-trees are defined as data structures, we can define operations such as adjunction and substitution and notions such as "derived structure". More precisely, we define quasi-trees to be structures consisting of pairs of nodes, called quasi-nodes, such that one is the "top" quasi-node and the other is the "bottom" quasi-node. The top and bottom quasi-node of a pair are linked by a dominance constraint. Bottom quasi-nodes immediately dominate top quasi-nodes of other quasi-node pairs, and each top quasi-node is immediately dominated by exactly one bottom quasi-node. For simplicity, we will assume that there is only a bottom root quasi-node (i.e., no top root quasi-node), and that bottom frontier quasi-nodes are omitted (i.e., frontier nodes just consist of top quasi-nodes). Furthermore, we will assume that each quasi-node has a label, and is equipped with a finite feature structure. A sample quasi-tree is shown in Figure 1 (quasi-tree $\alpha_5$ of Vijay-Shanker (1992, p.488)).

We follow Vijay-Shanker (1992, Section 2.5) in defining *substitution* as the operation of forming a quasi-node pair from a frontier node of one tree (which becomes the top node) and the root node of another tree (which becomes the bottom node). As always, a dominance link relates the two quasi-nodes of the newly formed pair. Adjunction is not defined separately: it suffices to say that a pair of quasi-nodes is "broken up", thus forming two quasi-trees. We then perform two substitutions. Observe that nothing keeps us from breaking up more than one pair of quasi-nodes in either of two quasi-trees, and then performing more than two substitutions (as long as dominance constraints are respected); there are no operations in regular TAG that correspond to such operations. We will say that a quasi-tree is *derived* if in all quasi-node pairs, the two quasi-nodes are equated, meaning that they have the same label and the two feature structures are unified, and furthermore, if all frontier quasi-nodes have terminal labels. The string associated with this quasi-tree is defined in the usual way.

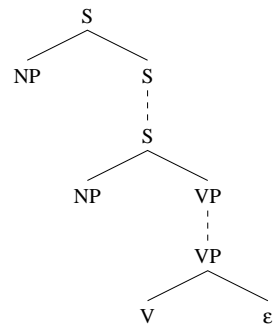

Figure 1: Sample quasi-tree

We have now fully defined a formalism (if informally): its data structures (quasi-trees), its combination operation (substitution), and the notion of derived structure. We will call this formalism Quasi-Tree Substitution Grammar (QTSG). It can easily be seen that all examples discussed by Vijay-Shanker (1992) are derivations in QTSG. The question arises as to the formal and computational properties of QTSG.

### Multiset-Valued LIG

In order to find a mathematical model for certain uses of multiset-valued feature structures, discussed above, we now introduce a multiset-valued variant of LIG. We denote by $\mathcal{M}(A)$ the set of multisets over the elements of $A$, and we use the standard set notation to refer to the corresponding multiset operations.

**Definition 1** *A* **multiset-valued Linear Index Grammar** *({}-LIG) is a 5-tuple* $(V_N, V_T, V_I, P, S)$, *where* $V_N$, $V_T$, *and* $V_I$ *are disjoint sets of terminals, non-terminals, and indices, respectively;* $S \in V_N$ *is the start symbol; and $P$ is a set of productions of the following form:*

$$p : As \longrightarrow v_0 B_1 s_1 v_1 \ldots v_{n-1} B_n s_n v_n$$

*for some* $n \geq 0$, $A, B_1, \ldots, B_n \in V_N$, $s, s_1, \ldots, s_n$ *multisets of members of* $V_I$, *and* $v_0, \ldots, v_n \in V_T^*$.

*The derivation relation* $\Longrightarrow$ *for a {}-LIG is defined as follows. Let* $\beta, \gamma \in (V_N \mathcal{M}(V_I) \cup V_T)^*$, $t, t_1, \ldots, t_n$ *multisets of members of* $V_I$, *and* $p \in P$ *of the form given above. Then we have*

$$\beta A t \gamma \stackrel{p}{\Longrightarrow} \beta v_0 B_1 t_1 v_1 \ldots v_{n-1} B_n t_n v_n \gamma$$

*such that* $t = \cup_{i=1}^{n}(t_i \setminus s_i) \cup s$. *If $G$ is a {}-LIG,* $L(G) = \{w \mid S \stackrel{*}{\Longrightarrow}_G w, w \in V_T^*\}$.

Suppose we want to apply rule $p$ to an instance of nonterminal $A$ with an index multiset $t$ in a sentential form. First, we remove the indices in $s$ from $t$, then we rewrite the nonterminal, then we distribute the remaining indices freely among the newly introduced nonterminals $B_1, \ldots, B_n$, creating new multisets, and finally we add $s_i$ to the new multiset for each $B_i$, creating the new $t_i$.

The reader will have noticed, and hopefully excused, the abuse of notation in this definition, which results from mixing set-notation and string-notation. We can also define {}-LIG as a pure string-rewriting system which does not require the definition of additional data structures (the multisets) for the notion of "derivation" (see (Rambow, 1994)). However, the definition provided here (using an explicit representation of multisets) has the advantage of corresponding more directly to the intuition underlying {}-LIG and is much easier to understand and use in proofs. The issue is purely notational.

We now introduce a restriction on derivations, which will be useful later.

**Definition 2** A **linearly-restricted** *derivation in a* {}-LIG *is a derivation* $\varrho : S \overset{*}{\Longrightarrow} w$ *with* $w \in V_T^*$ *such that:*

1. *The number of index symbols added (and hence removed) during the derivation is linearly bounded by* $|w|$.
2. *The number of $\varepsilon$-productions used during the derivation is linearly bounded by* $|w|$.

We let $L_R(G) = \{w \mid \text{there is a derivation } \varrho : S \overset{*}{\Longrightarrow} w \text{ such that } \varrho \text{ is linearly-restricted}\}$, and we let $\mathcal{L}_R(\{\}\text{-LIG}) = \{L_R(G) \mid G \text{ a } \{\}\text{-LIG}\}$. If $G$ is a {}-LIG such that $L_R(G) = L(G)$, we say that $G$ is linearly restricted. Many of the results that we will show apply only to linearly restricted {}-LIGs. However, as we will see, all linguistic applications will make use of this restricted version.

EXAMPLE 1
The following grammar derives the language COUNT-5, where COUNT-5 = $\{a^n b^n c^n d^n e^n \mid n \geq 0\}$.
Let $G_1 = (V_N, V_T, V_I, P, S)$ with:
$V_N = \{S, A, B, C, D, E\}$
$V_T = \{\text{a,b,c,d,e}\}$
$V_I = \{s_a, s_b, s_c, s_d, s_e\}$
$P = \{p_1 : S \longrightarrow S\{s_a, s_b, s_c, s_d, s_e\}$
$p_2 : S \longrightarrow ABCDE$
$p_3 : A\{s_a\} \longrightarrow Aa, \quad p_4 : A \longrightarrow \varepsilon$
$p_5 : B\{s_b\} \longrightarrow Bb, \quad p_6 : B \longrightarrow \varepsilon$
$p_7 : C\{s_c\} \longrightarrow Cc, \quad p_8 : C \longrightarrow \varepsilon$
$p_9 : D\{s_d\} \longrightarrow Dd, \quad p_{10} : D \longrightarrow \varepsilon$
$p_{11} : E\{s_e\} \longrightarrow Ee, \quad p_{12} : E \longrightarrow \varepsilon\ \}$
A sample derivation is shown in Figure 2.

This example shows that $\mathcal{L}(\{\}\text{-LIG})$ is not contained in $\mathcal{L}(\text{LIG})$, since the latter cannot derive COUNT-5. We now define two normal forms which will be used later. We omit the proofs and refer to (Rambow, 1994) for details.

**Definition 3** *A* {}-*LIG* $G = (V_N, V_T, V_I, P, S)$ *is in* **restricted index normal form** *or RINF if all productions in P are of one of the following forms (where $A, B \in V_N$, $f \in V_I$ and $\alpha \in (V_T \cup V_N)^*$):*

1. $A \longrightarrow \alpha$
2. $A \longrightarrow Bf$
3. $Af \longrightarrow B$

**Theorem 1** *For any* {}-*LIG, there is an equivalent* {}-*LIG in RINF.*

**Definition 4** *A* {}-*LIG* $G = (V_N, V_T, V_I, P, S)$ *is in* **Extended Two Form** *(ETF) if every production in P has the form* $As \rightarrow B_1 s_1 B_2 s_2$, $As \rightarrow Bs'$, *or* $A \rightarrow a$, *where* $A, B_1, B_2 \in V_N$, $s, s_1, s_2, s' \in V_I^*$, *and* $a \in V_T \cup \{\varepsilon\}$.

**Theorem 2** *For any* {}-*LIG, there is an equivalent* {}-*LIG in ETF.*

We now discuss some formal properties of {}-LIG. For reasons of space limitation, we only sketch the proofs; full versions can be found in (Rambow, 1994). We start with the weak generative power. We have already seen that {}-LIG can generate languages not in $\mathcal{L}(\text{LIG})$ (and hence not in $\mathcal{L}(\text{TAG})$). We will now show that linearly restricted {}-LIGs are at most context-sensitive.

**Theorem 3** $\mathcal{L}_R(\{\}\text{-LIG}) \subseteq \mathcal{L}(\text{CSG})$.

*Outline of the proof.* We simulate a derivation in a linear bounded automaton. The space needed for this is bounded linearly in the length of the input word, since the number of the symbols that are erased, the index symbols and nonterminals that rewrite to $\varepsilon$, is linearly bounded. ∎

What sort of languages could a {}-LIG possibly not generate? Consider the copy language $L = \{ww \mid w \in \{a, b\}^*\}$, and let us suppose that it is generated by $G$, a {}-LIG. This language cannot be generated by a CFG. We therefore know that for any integer $M$, there are infinitely many strings in $L$ whose derivation in $G$ is such that at some point, an index multiset in the sentential form contains more than $M$ index symbols (since any finite use of index symbols can be simulated by a pure CFG). It must be the case that this unbounded multiset is crucial in restricting the second half of the generated string in such a way that it copies the first half (again, since a pure CFG cannot derive such strings). However, it is impossible for a data structure like a (multi-)set (over a finite index alphabet) to record the required sequential information. Therefore, the second half of the string cannot be adequately constrained, and $G$ cannot exist. This argument motivates the following conjecture.

**Conjecture 4** $\{ww \mid w \in \{a, b\}^*\}$ *is not in* $\mathcal{L}(\{\}\text{-LIG})$.

$$S \stackrel{p_1}{\Longrightarrow} S\{s_a, s_b, s_c, s_d, s_e\}$$
$$\stackrel{p_1 p_1}{\Longrightarrow} S\{s_a, s_b, s_c, s_d, s_e, s_a, s_b, s_c, s_d, s_e, s_a, s_b, s_c, s_d, s_e\}$$
$$\stackrel{p_2}{\Longrightarrow} A\{s_a, s_a, s_a\} B\{s_b, s_b, s_b\} C\{s_c, s_c, s_c\} D\{s_d, s_d, s_d\} E\{s_e, s_e, s_e\}$$
$$\stackrel{p_7}{\Longrightarrow} A\{s_a, s_a, s_a\} B\{s_b, s_b, s_b\} C\{s_c, s_c\} cD\{s_d, s_d, s_d\} E\{s_e, s_e, s_e\}$$
$$\stackrel{p_3 p_3 p_4}{\Longrightarrow} aaaB\{s_b, s_b, s_b\} C\{s_c, s_c\} cD\{s_d, s_d, s_d\} E\{s_e, s_e, s_e\}$$
$$\stackrel{*}{\Longrightarrow} aaabbbcccdddeee$$

Figure 2: Sample derivation in {}-LIG $G_1$

We now turn to closure properties.

**Theorem 5** $\mathcal{L}(\{\}\text{-LIG})$ *is a substitution-closed full abstract family of languages (AFL).*

*Outline of the proof.* Since $\mathcal{L}(\{\}\text{-LIG})$ contains all context-free languages, it contains all regular languages, and therefore it is sufficient to show that $\mathcal{L}(\{\}\text{-LIG})$ is closed under intersection with regular languages and substitution. These results are shown by adapting the techniques used to show the corresponding results for CFGs. ∎

Finally, we turn to the recognition and parsing problem. Again, we will restrict our attention to the linearly restricted version of {}-LIG.

**Theorem 6** *Each language in $\mathcal{L}_R(\{\}\text{-LIG})$ can be recognized in polynomial deterministic time.*

*Outline of the proof.* We extend the CKY parser for CFG. Let $G$ be a {}-LIG in ETF. Since $G$ may contain $\varepsilon$-productions, the algorithm is adapted by letting the indices of the matrix refer to positions between symbols in the input string, not the symbols themselves. In order to account for the index multiset, we let the entries in the recognition matrix be pairs consisting of a nonterminal symbol and a $|V_I|$-tuple of integers:

$$\langle A, (n_1, \ldots, n_{|V_I|}) \rangle$$

The $|V_I|$-tuple of integers represents a multiset, with each integer designating the number of copies of a given index symbol that the set contains. In an entry of the matrix, each pair represents a partial derivation of a substring of the input string. More precisely, if the input word is $a_1 \cdots a_n$, and if $V_I = \{i_1, \ldots, i_{|V_I|}\}$, then we have $\langle A, (n_1, \ldots, n_{|V_I|}) \rangle$ in entry $t_{i,j}$ of the recognition matrix if and only if there is a derivation $As \Longrightarrow a_{i+1} \cdots a_j$, where multiset $s$ contains $n_k$ copies of index symbol $i_k$, $1 \leq k \leq |V_I|$. Clearly, there is a derivation in the grammar if and only if entry $t_{0,n}$ contains the pair $\langle S, (0, \ldots, 0) \rangle$. Now since the grammar is linearly restricted, each $n_k$ is bounded by $n$, and hence the number of different pairs is linearly bounded by $|V_N| n^{|V_I|}$. Thus each entry in the matrix can be computed in $O(n^{1+2|V_I|})$ steps, and since there are $O(n^2)$ entries, we get an overall time complexity of $O(n^{3+2|V_I|})$. ∎

## UVG with Dominance Links

We now formally define UVG with dominance links (UVG-DL), which serves as a formal model for the second and third phenomena introduced above, word order variation and quasi-trees. The definition differs from that of UVG only in that vectors are equipped with dominance relations which impose an additional condition on derivations. Note that the definition refers to the notion of derivation tree of a UVG, which is defined as for CFG.

**Definition 5** *An* **Unordered Vector Grammar with Dominance Links** *(UVG-DL) is a 4-tuple $(V_N, V_T, V, S)$, where $V_N$ and $V_T$ are sets of nonterminals and terminals, respectively, $S$ is the start symbol, and $V$ is a set of vectors of context-free productions equipped with dominance links. For a given vector $v \in V$, the dominance links form a binary relation $\mathsf{dom}_v$ over the set of occurrences of non-terminals in the productions of $v$ such that if $\mathsf{dom}_v(A, B)$, then $A$ (an instance of a symbol) occurs in the right-hand side of some production in $v$, and $B$ is the left-hand symbol (instance) of some production in $v$.*

*If $G$ is a UVG-DL, $L(G)$ consists of all words $w \in V_T^*$ which have a derivation $\varrho$ of the form*

$$S \stackrel{p_1}{\Longrightarrow} w_1 \stackrel{p_2}{\Longrightarrow} w_2 \ldots w_{r-1} \stackrel{p_r}{\Longrightarrow} w_r = w,$$

*such that $\varrho$ meets the following two conditions:*

1. *$p_1 p_2 \ldots p_r$ is a permutation of a member of $V^*$.*
2. *The dominance relations of $V$, when interpreted as the standard dominance relation defined on trees, hold in the derivation tree of $\varrho$.*

The second condition can be formulated as follows: if $v$ in $V$ contributes instances of productions $p_1$ and $p_2$ (and perhaps others), and the $k$th daughter in the right-hand side of $p_1$ dominates the left-hand nonterminal of $p_2$, then in the context-free derivation tree associated with $\varrho$ (the unique node associated with) the $k$th daughter node of $p_1$ dominates (the unique node associated with) $p_2$. We now give an example. (The superscripts distinguish instances of symbols and are not part of the nonterminal alphabet.)

EXAMPLE 2

Let $G_2 = (V_N, V_T, V, S')$ with:

$v_1$: $\{(S' \longrightarrow da\beta\ VP)\}$ with $\mathsf{dom}_{v_1} = \emptyset$
$v_2$: $\{(VP^{(1)} \longrightarrow NP_{\text{nom}}\ VP^{(2)}), (VP^{(3)} \longrightarrow NP_{\text{dat}}\ VP^{(4)}), (VP^{(5)} \longrightarrow VP^{(6)}\ VP^{(7)}), (VP^{(8)} \longrightarrow verspricht)\}$ with $\mathsf{dom}_{v_2} = \{(VP^{(2)}, VP^{(8)}), (VP^{(4)}, VP^{(8)}), (VP^{(7)}, VP^{(8)})\}$
$v_3$: $\{(VP^{(1)} \longrightarrow VP^{(1)}\ VP^{(2)}), (VP^{(3)} \longrightarrow zu\ versuchen)\}$ with $\mathsf{dom}_{v_3} = \{(VP^{(2)}, VP^{(3)})\}$
$v_4$: $\{(VP^{(1)} \longrightarrow NP_{\text{acc}}\ VP^{(2)}), (VP^{(3)} \longrightarrow zu\ reparieren)\}$ with $\mathsf{dom}_{v_4} = \{(VP^{(2)}, VP^{(3)})\}$
$v_5$: $\{(NP_{\text{nom}} \longrightarrow der\ Meister)\}$ with $\mathsf{dom}_{v_5} = \emptyset$
$v_6$: $\{(NP_{\text{dat}} \longrightarrow niemandem)\}$ with $\mathsf{dom}_{v_6} = \emptyset$
$v_7$: $\{(NP_{\text{acc}} \longrightarrow den\ K\ddot{u}hlschrank)\}$ with $\mathsf{dom}_{v_7} = \emptyset$

Figure 3: Definition of $V$ for UVG-DL $G_2$

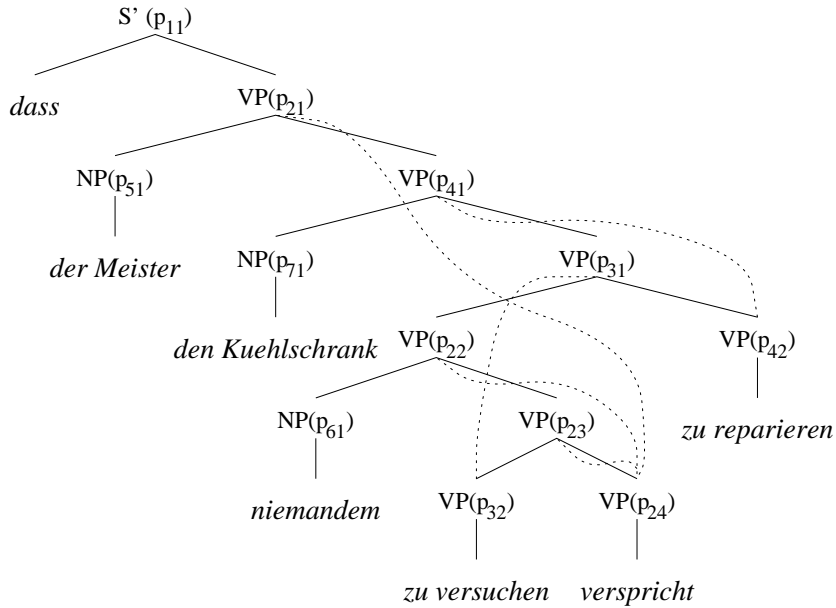

Figure 4: Sample UVG-DL derivation

$V_N = \{S', VP, NP_{\text{nom}}, NP_{\text{dat}}, NP_{\text{acc}}\}$
$V_T = \{da\beta, verspricht, zu\ versuchen, zu\ reparieren, der\ Meister, niemandem, den\ K\ddot{u}hlschrank\}$[2]
$V = \{v_1, v_2, v_3, v_4, v_5, v_6, v_7\}$

where the $v_i$ are as defined in Figure 3.

A sample derivation is shown in Figure 4, where the dominance relations are shown by dotted lines. Observe that the example grammar is lexicalized. We will denote the class of lexicalized UVG-DL by UVG-DL$_{\text{Lex}}$.

It is clear that the dominance links of UVG-DL are the additional constraints that we argued above are necessary to adequately restrict the structural relation between arguments and their verbs. Furthermore, UVG-DL is a notational variant of QTSG: every vector represents a quasi-tree, and identifying quasi-nodes corresponds to rewriting. The condition on a successful derivation in QTSG – that all nonterminal nodes be identified – corresponds to the definition of a derivation in UVG-DL. We have therefore found a mathematical model for the second and third phenomenon mentioned in Section 2.

We now turn to the formal properties of UVG-DL. Our main result is that UVG-DL is weakly equivalent to $\{\}$-LIG. The sets of a $\{\}$-LIG implement the dominance links and make sure that all members from one set of rules are used during a derivation. We first introduce some more terminology with which to describe the derivations of UVG-DLs. If two productions $p_{v,1}$ and $p_{v,2}$ from vector $v$ are linked by a dominance link from a right-hand side nonterminal of $p_{v,1}$ to the left-hand nonterminal $p_{v,2}$, then we will denote this link by $l_{v,1,2}$. We will say that $p_{v,1}$ (or the right-hand side nonterminal in question) has a *passive dominance requirement* of $l_{v,1,2}$, and that $p_{v,2}$ has an *active dominance requirement* of $l_{v,1,2}$. If $p_{v,1}$ or $p_{v,2}$ is used in a partial derivation such that the other production is not used in the derivation, the dominance requirement (passive or active) will be called *unfulfilled*. Let $\varrho$ be a (partial) derivation. We associate with $\varrho$ a multi-set which represent all the unfulfilled active dominance requirements of $\varrho$, written $\top(\varrho)$.

**Theorem 7** $\mathcal{L}(\text{UVG-DL}) = \mathcal{L}(\{\}\text{-LIG})$

---

[2]Gloss (in order): that, promises, to try, to repair, the master, no-one, the refrigerator.

*Outline of the proof.* The theorem is proved in two parts (one for each inclusion). We first show the inclusion $\mathcal{L}(\text{UVG-DL}) \subseteq \mathcal{L}(\{\}\text{-LIG})$. Let $G = (V_N, V_T, V, S)$ be a UVG-DL, where $V = \{v_1, \ldots, v_K\}$ with $v_i = (p_{i,1}, \ldots, p_{i,k_i})$, $k_i = |v_i|$, $1 \leq i \leq K$. We construct a $\{\}$-LIG $G' = (V_N, V_T, V_I, P, S)$. Let $V_I = \{l_{i,j,k} \mid 1 \leq i \leq K, 1 \leq j, k \leq k_i\}$. Define $P$ as follows.

Let $v$ in $V$, and let $p$ in $v$ be the production $A \longrightarrow w_0 B_1 w_1 \cdots B_n w_n$ be in $v_r$. In the following, we will denote by $\top(p)$ the multiset of active dominance requirements of $p$, and by $\bot_i(p)$ the multiset of passive dominance requirements of $B_i$, $1 \leq i \leq n$. Add to $P$ the following production:

$$A \top(p) \longrightarrow w_0 B_1 \bot_1(p) w_1 \cdots B_n \bot_n(p) w_n$$

$P$ contains no other productions. We show by induction that for $A$ in $V_N$, and $w$ in $V_T^*$, we have $A \stackrel{*}{\Longrightarrow}_G w$ iff $A \stackrel{*}{\Longrightarrow}_{G'} w$. Specifically, we show that for all integers $k$, $\varrho : A \stackrel{k}{\Longrightarrow}_G w$, $w \in V_T^*$, with unfulfilled active dominance requirements $\top(\varrho)$, implies that there is a derivation $A\top(\varrho) \stackrel{*}{\Longrightarrow}_{G'} w$, and, conversely, we show that for all integers $k$, $At \stackrel{i}{\Longrightarrow}_{G'} \alpha$, $A \in V_N$, $t$ a multiset of elements of $V_I$, and $\alpha \in V_T^*$, implies that there is a derivation $\varrho : A \stackrel{*}{\Longrightarrow}_G \alpha$ such that $\top(\varrho) = t$.

For the inclusion $\mathcal{L}(\{\}\text{-LIG}) \subseteq \mathcal{L}(\text{UVG-DL})$, we take a slightly different approach to avoid notational complexity. Let $G = (V_N, V_T, V_I, P, S)$ be a $\{\}$-LIG in RINF. We construct a UVG-DL $G' = (V_N, V_T, V, S)$, where $V$ is defined as follows:

1. If $p \in P$ is a $\{\}$-LIG production of RINF type 1, then $((p), \emptyset) \in V$.

2. If $p \in P$ is a $\{\}$-LIG production of RINF type 2, with $p = A \longrightarrow Bf$ for $A, B \in V_N$, $f \in V_I$, then for all $q \in P$ such that $q = Cf \longrightarrow D$, $v = ((A \longrightarrow B, C \longrightarrow D), \text{dom}_v(B, C))$ is in $V$.

Let $A$ be in $V_N$, and $w$ in $V_T^*$. We show by induction that $S \stackrel{*}{\Longrightarrow}_G w$ iff $S \stackrel{*}{\Longrightarrow}_{G'} w$. Specifically, we first show that for all integers $k$, for all $\{\}$-LIGs $G$ and the corresponding UVG-DL $G'$ as constructed above, if there is a derivation $\varrho : S\{\} \stackrel{*}{\Longrightarrow}_G w$ with $k$ instances of applications of rules of type 2, then there is a derivation $\varrho' : S \stackrel{*}{\Longrightarrow}_{G'} w$ such that $\varrho$ and $\varrho'$ are identical except for the index symbols in the sentential forms of $\varrho$. For the converse inclusion, we show that for all integers $k$, for all $\{\}$-LIGs $G$ and the correspond UVG-DL $G'$ as constructed above, if there is a derivation $\varrho' : S\{\} \stackrel{*}{\Longrightarrow}_{G'} w$ with $k$ instances of applications of rules from vectors with two elements, then there is a derivation $\varrho : S \stackrel{*}{\Longrightarrow}_G w$ such that $\varrho$ and $\varrho'$ are identical except for the index symbols in the sentential forms of $\varrho$. ∎

This equivalence lets us transfer results from $\{\}$-LIG to UVG-DL. It can easily be seen from the construction employed in the proof of Theorem 7 that a lexicalized UVG-DL maps to a linearly restricted $\{\}$-LIG. For linguistic purposes we are only interested in lexicalized grammars, and therefore the linear restriction is quite natural. We obtain the following corollaries thanks to Theorem 7.

**Corollary 8** $\mathcal{L}(\text{UVG-DL}_{\text{Lex}}) \subseteq \mathcal{L}(\text{CSG})$.

**Corollary 9** $\mathcal{L}(\text{UVG-DL})$ *is a substitution-closed full AFL.*

**Corollary 10** *Each language in $\mathcal{L}(\text{UVG-DL}_{\text{Lex}})$ can be recognized in polynomial deterministic time.*

## Related Formalisms

Based on word-order facts from Turkish, Hoffman (1992) proposes an extension to CCG called $\{\}$-CCG, in which arguments of functors form sets, rather than being represented in a curried notation. Under function composition, these sets are unioned. Thus the move from CCG to $\{\}$-CCG corresponds very much to the move from LIG to $\{\}$-LIG. We conjecture that (a version of) $\{\}$-CCG is weakly equivalent to $\{\}$-LIG.

Staudacher (1993) defines a related system called *distributed index grammar* or DIG. DIG is like LIG, except that the stack of index symbols can be split into chunks and distributed among the daughter nodes. However, the formalism is not convincingly motivated by the linguistic data given (which can also be handled by a simple LIG) or by other considerations.

Several extensions to $\{\}$-LIG and UVG-DL are defined in (Rambow, 1994). First, we can introduce the "integrity" constraint suggested by Becker et al. (1991) which restricts long-distance relations through nodes. This is necessary to implement the linguistic notion of "barrier" or "island". Second, we can define the tree-rewriting version of UVG-DL, called V-TAG. This is motivated by Conjecture 4, which (if true) means that UVG-DL cannot derive Swiss German. Under either extension, the weak generative power is extended, but the formal and computational results obtained for $\{\}$-LIG and UVG-DL still hold.

## Conclusion

This paper has presented two equivalent formalisms, $\{\}$-LIG and UVG-DL, which provide formal models for the three different phenomena that we identified in the beginning of the paper. We have shown that both formalisms, under certain restrictions that are compatible with the motivating phenomena, are restricted in their generative capacity and polynomially parsable, thus making them attractive candidates for modeling natural language. Furthermore, the formalisms are substitution-closed AFLs, suggesting that the definitions we have given are "natural" from the point of view of formal language theory.

## Acknowledgments

I would like to thank Bob Kasper, Gaëlle Recourcé, Giorgio Satta, Ed Stabler, two anonymous reviewers,

and especially K. Vijay-Shanker for useful comments and discussions. The research reported in this paper was conducted while the author was with the Computer and Information Science Department of the University of Pennsylvania. The research was sponsored by the following grants: ARO DAAL 03-89-C-0031; DARPA N00014-90-J-1863; NSF IRI 90-16592; and Ben Franklin 91S.3078C-1.